\documentclass{article}

\pdfoutput=1

    \usepackage[final,nonatbib]{neurips_2021}

\usepackage[utf8]{inputenc} 
\usepackage[T1]{fontenc}    
\usepackage{hyperref}       
\usepackage{url}            
\usepackage{booktabs}       
\usepackage{amsfonts}       
\usepackage{nicefrac}       
\usepackage{microtype}      
\usepackage{xcolor}         
\usepackage{amsmath}

\usepackage{physics}
\usepackage{mathdots}
\usepackage{yhmath}
\usepackage{cancel}
\usepackage{color}
\usepackage{siunitx}
\usepackage{array}
\usepackage{multirow}
\usepackage{amssymb}
\usepackage{caption}
\usepackage{subcaption}
\usepackage{stmaryrd}
\usepackage{esvect}
\usepackage{bm}
\usepackage{graphicx}

\usepackage{physics}
\usepackage{amsmath}
\usepackage{tikz}
\usepackage{mathdots}
\usepackage{yhmath}
\usepackage{cancel}
\usepackage{color}
\usepackage{siunitx}
\usepackage{array}
\usepackage{multirow}
\usepackage{amssymb}
\usepackage{gensymb}
\usepackage{tabularx}
\usepackage{extarrows}
\usepackage{booktabs}
\usetikzlibrary{fadings}
\usetikzlibrary{patterns}
\usetikzlibrary{shadows.blur}
\usetikzlibrary{shapes}

\definecolor{bluecite}{HTML}{0875b7}
\hypersetup{
  colorlinks=true,
  citecolor=bluecite,
  linkcolor=magenta
}

\title{One Step at a Time: Pros and Cons of Multi-Step Meta-Gradient Reinforcement Learning}

\author{
  Clément Bonnet\thanks{Correspondence to: <\texttt{c.bonnet@instadeep.com}>}\\
  InstaDeep\\
  \and
  \textbf{Paul Caron}\\
  InstaDeep\\
  \and
  \textbf{Thomas Barrett}\\
  InstaDeep\\
  \and
  \textbf{Ian Davies}\\
  InstaDeep\\
  \and
  \textbf{Alexandre Laterre}\\
  InstaDeep\\
}

\begin{document}

\maketitle

\begin{abstract}
Self-tuning algorithms that adapt the learning process online encourage more effective and robust learning. Among all the methods available, meta-gradients have emerged as a promising approach. They leverage the differentiability of the learning rule with respect to some hyper-parameters to adapt them in an online fashion. Although meta-gradients can be accumulated over multiple learning steps to avoid myopic updates, this is rarely used in practice. In this work, we demonstrate that whilst multi-step meta-gradients do provide a better learning signal in expectation, this comes at the cost of a significant increase in variance, hindering performance. In the light of this analysis, we introduce a novel method mixing multiple inner steps that enjoys a more accurate and robust meta-gradient signal, essentially trading off bias and variance in meta-gradient estimation. When applied to the Snake game, the mixing meta-gradient algorithm can cut the variance by a factor of 3 while achieving similar or higher performance.
\end{abstract}

\section{Introduction}

Over the past decade, the most successful machine learning applications have moved from handcrafted features, subject to the oversights and biases of the designer, to systems where representations are learned from raw input data in an end-to-end manner. However, the algorithms with which this learning is realised have remained manually designed. Recently, meta-learning has emerged with the potential of systems that adapt to the specific learning task, promising out-of-distribution generalisation and robust learning.

In this work, we consider the application of meta-learning to reinforcement learning (RL), so-called meta-RL. By considering the “lifetime” of an agent to be training the agent on a specific task, we can distinguish two main frameworks within meta-RL: “multi-” and “single-” lifetime~\cite{xu2020metagradient}. The multi-lifetime setting considers an algorithm that is given a set of training tasks. The aim is to quickly adapt to a new task either by learning an initialisation \cite{finn2017modelagnostic, gupta2018metareinforcement, nichol2018firstorder, rakelly2019efficient, zintgraf2019fast}, or by learning an objective \cite{bechtle2021metalearning, houthooft2018evolved, kirsch2020improving, oh2021discovering}. By contrast, single-lifetime optimisation aims to modify the learning process online to improve performance on the task. The algorithm is tuned as the system is being optimised to solve a problem. In RL in particular, it has been shown that hyper-parameters have a huge impact on performance and training stability~\cite{amit2020discount,BertsekasTsitsiklis96}. A wide variety of algorithms for self-tuning have been proposed, based on the works of Xu et al. (2018)~\cite{xu2018metagradient} and Zheng et al. (2018)~\cite{zheng2018learning}, which use meta-gradients to update hyper-parameters online \cite{veeriah2019discovery, wang2020exponentially, xu2020metagradient, zahavy2021selftuning}.

Our efforts focus on meta-gradient RL~\cite{xu2018metagradient} in general from which many self-tuning algorithms have been designed, making our work applicable to a wide range of meta-gradient methods \cite{ veeriah2019discovery, wang2020exponentially,
xu2020metagradient,zahavy2021selftuning,
zheng2018learning,
zhou2020online}. Meta-gradients are used to learn and adapt differentiable components of the update rule such as the discount factor $\gamma$, or the bootstrapping factor $\lambda$ from the $\lambda$-return~\cite{schulman2018highdimensional}, or even the whole target parameterised by a neural network~\cite{xu2020metagradient}. In each of these cases, meta-gradients are computed by backpropagating through the inner update. Consequently, meta-parameters keep evolving to make the agent learn a bit better at each new step.



Although online meta-gradient updates have been widely adopted for self-tuning learning algorithms, they suffer from a poor signal-to-noise ratio, i.e. high variance, as well as a bias~\cite{flennerhag2021bootstrapped, metz2019understanding, wu2018understanding} arising from the very local and myopic properties of the meta-update. We focus on identifying the bias that arises from computing meta-gradients through an inner update.
Despite multi-step meta-gradients being more accurate in expectation, we show that they tend
to yield higher variance, which has limited its usage in practice.
We, therefore, demonstrate the need for and potential of trading off bias and variance in the way meta-gradients are computed. To address this issue, we propose a solution that mixes multiple inner updates to strongly reduce the variance, making meta-updates more robust with equal or better performance than multi-step meta-gradients for the same number of steps.


\section{Background}

In the context of deep reinforcement learning, both the value function $v$ and the policy $\pi$ are parameterised by neural networks with parameters $\theta$. They are trained using an RL objective whose hyper-parameters are denoted $\zeta$.
Self-tuning methods aim to adapt online some of these hyper-parameters, called \textit{meta-parameters} $\eta \subseteq \zeta$, such as to optimise the learning of the agent.

To study the bias and variance of meta-gradient methods, we first work on the problem of state-value prediction and then validate our findings on a control task using an actor-critic algorithm.
In both cases, the agent is trained using an inner loss $\mathcal{L}$, while the meta-learning agent updates its meta-parameters using an outer loss $\mathcal{L}'$. We also denote by $\theta'$ the parameters after one or several inner updates, $\tau'$ the trajectories sampled from $\pi_{\theta'}$, and $\eta'$ the fixed hyper-parameters of the meta-RL algorithm, representing a good proxy of the agent's final objective.

\subsection{Meta-Gradients for Value Prediction}
Similarly to Xu et al. (2018)~\cite{xu2018metagradient}, we first analyse the properties of meta-gradients in the context of value prediction using the TD($\lambda$) algorithm~\cite{sutton2018reinforcement}. In that context, the inner loss $\mathcal{L}$, whose gradient is presented below, is the Mean-Squared Error (MSE) between the value prediction $v_\theta(\tau)$ and the $\lambda$-return $g_\eta(\tau)$.
\begin{equation}
    \frac{\partial \mathcal{L}(\theta, \eta, \tau)}{\partial \theta} = -\left(g_\eta(\tau) - v_\theta(\tau)\right)\frac{\partial v_\theta(\tau)}{\partial \theta}
\end{equation}

The gradient of the outer loss $\mathcal{L}'$ is also constructed from the MSE loss but this time with respect to the meta-parameters $\eta$ and taken after one or several inner updates.
\begin{equation}
    \frac{\partial \mathcal{L}' (\theta', \eta', \tau')}{\partial \eta} = -\left(g_{\eta'}(\tau') - v_{\theta'}(\tau')\right)\frac{\partial v_{\theta'}(\tau')}{\partial \eta}
\end{equation}

\subsection{Meta-Gradients for Control}
For control, we focus on the A2C algorithm~\cite{Konda00actor-criticalgorithms}
in which an actor updates the policy distribution in the direction suggested by the critic whose goal is to estimate the state-value function.
Both the value function $v$ and the policy $\pi$ are parameterised by neural networks with parameters $\theta$, and are trained using the A2C objective whose hyper-parameters are denoted $\eta$. Self-tuning aims to adapt $\eta$ online such as to optimise the learning of the agent. Any subset of differentiable hyper-parameters can be tuned using meta-gradients~\cite{zahavy2021selftuning}. Here, we consider $\eta = \{\gamma, \lambda, c_{\textrm{crit}}, c_{\textrm{entr}}\}$, meaning we allow the agent to meta-learn all the differentiable hyper-parameters of the A2C objective.

The gradient of the inner loss $\mathcal{L}$ constructed from the A2C objective is given as follows.
\begin{equation}
    \frac{\partial \mathcal{L}(\theta, \eta, \tau)}{\partial \theta} = -(g_\eta(\tau) - v_\theta(\tau))\frac{\partial \log \pi_\theta(\tau)}{\partial \theta} - c_{\textrm{crit}} (g_\eta(\tau) - v_\theta(\tau)) \frac{\partial v_\theta(\tau)}{\partial \theta} - c_{\textrm{entr}}\frac{\partial H(\pi_\theta(\tau))}{\partial \theta}
\end{equation}

The inner loss combines the policy gradient term which aims to update the policy, with the MSE loss of the critic predictions and entropy regularisation. The outer loss results directly from the policy gradient theorem applied to the updated policy.

\begin{equation}
    \frac{\partial \mathcal{L}' (\theta', \eta', \tau')}{\partial \eta} = -\left(g_{\eta'}\left(\tau'\right) - v_{\theta'}\left(\tau'\right)\right)\frac{\partial \log \pi_{\theta'}(\tau')}{\partial \eta}
\end{equation}


\subsection{Meta-Gradients}

\begin{figure}[t]
    \centering
    \tikzset{every picture/.style={line width=0.75pt}} 

\begin{tikzpicture}[x=0.75pt,y=0.75pt,yscale=-1,xscale=1]

\draw   (82,43) .. controls (82,31.95) and (90.95,23) .. (102,23) .. controls (113.05,23) and (122,31.95) .. (122,43) .. controls (122,54.05) and (113.05,63) .. (102,63) .. controls (90.95,63) and (82,54.05) .. (82,43) -- cycle ;

\draw   (82,100) -- (117,100) -- (117,135) -- (82,135) -- cycle ;
\draw    (131,43) -- (209,43) ;
\draw [shift={(211,43)}, rotate = 180] [fill={rgb, 255:red, 0; green, 0; blue, 0 }  ][line width=0.08]  [draw opacity=0] (12,-3) -- (0,0) -- (12,3) -- cycle    ;
\draw    (408.2,43.36) -- (486.5,44.55) ;
\draw [shift={(488.5,44.58)}, rotate = 180.87] [fill={rgb, 255:red, 0; green, 0; blue, 0 }  ][line width=0.08]  [draw opacity=0] (12,-3) -- (0,0) -- (12,3) -- cycle    ;
\draw  [dash pattern={on 0.84pt off 2.51pt}]  (351.37,43.36) -- (408.25,43.36) ;
\draw [color={rgb, 255:red, 208; green, 2; blue, 27 }  ,draw opacity=1 ] [dash pattern={on 4.5pt off 4.5pt}]  (183.2,91.86) .. controls (235.08,68.88) and (443.48,92.47) .. (499.42,65.15) ;
\draw [shift={(501.87,63.86)}, rotate = 510.57] [fill={rgb, 255:red, 208; green, 2; blue, 27 }  ,fill opacity=1 ][line width=0.08]  [draw opacity=0] (10.72,-5.15) -- (0,0) -- (10.72,5.15) -- (7.12,0) -- cycle    ;
\draw [color={rgb, 255:red, 208; green, 2; blue, 27 }  ,draw opacity=1 ] [dash pattern={on 4.5pt off 4.5pt}]  (487,58) -- (409,58) ;
\draw [shift={(407,58)}, rotate = 360] [fill={rgb, 255:red, 208; green, 2; blue, 27 }  ,fill opacity=1 ][line width=0.08]  [draw opacity=0] (12,-3) -- (0,0) -- (12,3) -- cycle    ;
\draw [color={rgb, 255:red, 208; green, 2; blue, 27 }  ,draw opacity=1 ]   (414,124) -- (440,124) ;
\draw [shift={(442,124)}, rotate = 180] [fill={rgb, 255:red, 208; green, 2; blue, 27 }  ,fill opacity=1 ][line width=0.08]  [draw opacity=0] (12,-3) -- (0,0) -- (12,3) -- cycle    ;
\draw    (414,108) -- (439.98,108) ;
\draw [shift={(441.98,108)}, rotate = 180] [fill={rgb, 255:red, 0; green, 0; blue, 0 }  ][line width=0.08]  [draw opacity=0] (12,-3) -- (0,0) -- (12,3) -- cycle    ;
\draw   (408,98) -- (527.53,98) -- (527.53,152.53) -- (408,152.53) -- cycle ;
\draw   (221,43) .. controls (221,31.95) and (229.95,23) .. (241,23) .. controls (252.05,23) and (261,31.95) .. (261,43) .. controls (261,54.05) and (252.05,63) .. (241,63) .. controls (229.95,63) and (221,54.05) .. (221,43) -- cycle ;

\draw   (499,43) .. controls (499,31.95) and (507.95,23) .. (519,23) .. controls (530.05,23) and (539,31.95) .. (539,43) .. controls (539,54.05) and (530.05,63) .. (519,63) .. controls (507.95,63) and (499,54.05) .. (499,43) -- cycle ;

\draw   (224,100) -- (259,100) -- (259,135) -- (224,135) -- cycle ;
\draw    (270,43) -- (348,43) ;
\draw [shift={(350,43)}, rotate = 180] [fill={rgb, 255:red, 0; green, 0; blue, 0 }  ][line width=0.08]  [draw opacity=0] (12,-3) -- (0,0) -- (12,3) -- cycle    ;
\draw [color={rgb, 255:red, 208; green, 2; blue, 27 }  ,draw opacity=1 ] [dash pattern={on 4.5pt off 4.5pt}]  (349,58) -- (271,58) ;
\draw [shift={(269,58)}, rotate = 360] [fill={rgb, 255:red, 208; green, 2; blue, 27 }  ,fill opacity=1 ][line width=0.08]  [draw opacity=0] (12,-3) -- (0,0) -- (12,3) -- cycle    ;
\draw [color={rgb, 255:red, 208; green, 2; blue, 27 }  ,draw opacity=1 ] [dash pattern={on 4.5pt off 4.5pt}]  (210,58) -- (132,58) ;
\draw [shift={(130,58)}, rotate = 360] [fill={rgb, 255:red, 208; green, 2; blue, 27 }  ,fill opacity=1 ][line width=0.08]  [draw opacity=0] (12,-3) -- (0,0) -- (12,3) -- cycle    ;
\draw [color={rgb, 255:red, 208; green, 2; blue, 27 }  ,draw opacity=1 ]   (131,120) -- (209,120) ;
\draw [shift={(211,120)}, rotate = 180] [fill={rgb, 255:red, 208; green, 2; blue, 27 }  ,fill opacity=1 ][line width=0.08]  [draw opacity=0] (12,-3) -- (0,0) -- (12,3) -- cycle    ;
\draw [color={rgb, 255:red, 208; green, 2; blue, 27 }  ,draw opacity=1 ] [dash pattern={on 4.5pt off 4.5pt}]  (414,141) -- (440,141) ;
\draw [shift={(442,141)}, rotate = 180] [fill={rgb, 255:red, 208; green, 2; blue, 27 }  ,fill opacity=1 ][line width=0.08]  [draw opacity=0] (12,-3) -- (0,0) -- (12,3) -- cycle    ;

\draw (169.55,32.64) node  [font=\scriptsize] [align=left] {\begin{minipage}[lt]{52.43pt}\setlength\topsep{0pt}
$\displaystyle \nabla _{\theta }\mathcal{L}( \theta _{t} ,\ \eta _{t})$
\end{minipage}};
\draw (308.45,32.3) node  [font=\scriptsize] [align=left] {\begin{minipage}[lt]{56.37pt}\setlength\topsep{0pt}
$\displaystyle \nabla _{\theta }\mathcal{L}\left( \theta _{t}^{( 1)} ,\ \eta _{t}\right)$
\end{minipage}};
\draw (450.24,32.3) node  [font=\scriptsize] [align=left] {\begin{minipage}[lt]{62.24pt}\setlength\topsep{0pt}
$\displaystyle \nabla _{\theta }\mathcal{L}\left( \theta _{t}^{( n-1)} ,\ \eta _{t}\right)$
\end{minipage}};
\draw (168.43,108.44) node  [font=\scriptsize,color={rgb, 255:red, 208; green, 2; blue, 27 }  ,opacity=1 ] [align=left] {\begin{minipage}[lt]{50.9pt}\setlength\topsep{0pt}
$\displaystyle \nabla _{\eta }\mathcal{L} '\left( \theta _{t}^{( n)}( \eta _{t})\right)$
\end{minipage}};
\draw (95.53,35.5) node [anchor=north west][inner sep=0.75pt]  [font=\normalsize] [align=left] {$\displaystyle \theta _{t}$};
\draw (229.46,33.5) node [anchor=north west][inner sep=0.75pt]  [font=\normalsize] [align=left] {$\displaystyle \theta _{t}^{( 1)}$};
\draw (508.15,33.5) node [anchor=north west][inner sep=0.75pt]  [font=\normalsize] [align=left] {$\displaystyle \theta _{t}^{( n)}$};
\draw (449,117) node [anchor=north west][inner sep=0.75pt]  [font=\footnotesize] [align=left] {{\small Meta Update}};
\draw (449,100.5) node [anchor=north west][inner sep=0.75pt]  [font=\footnotesize] [align=left] {{\small Inner Update}};
\draw (91.41,113) node [anchor=north west][inner sep=0.75pt]  [font=\normalsize] [align=left] {$\displaystyle \eta _{t}$};
\draw (226.76,113) node [anchor=north west][inner sep=0.75pt]  [font=\normalsize] [align=left] {$\displaystyle {\displaystyle \eta _{t+1}}$};
\draw (449,134) node [anchor=north west][inner sep=0.75pt]  [font=\footnotesize] [align=left] {{\small Meta-Gradient}};

\end{tikzpicture}
    \caption{$n$-step meta-gradient estimation. Starting from $\theta_t$, the algorithm makes $n$ inner updates to get $\theta_t^{(n)}$, using the same meta-parameters $\eta_t$. The $n$-step meta-gradient 
    is then computed by backpropagating through the $n$ gradient steps. $\theta_{t+1}$ is updated to $\theta_t^{(1)}$ and the other $\theta_t^{(i)}, i>1$ are discarded. The process is then repeated starting from $\theta_{t+1}$ using meta-parameters $\eta_{t+1}$.
    and so on.
    }
    \label{fig:meta_gradient_diagram}
\end{figure}
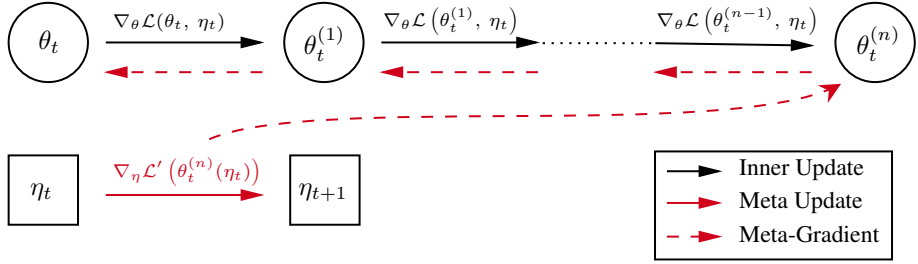

After one or multiple inner updates, one can estimate the meta-gradient by differentiating through the inner update(s) to minimise an outer loss $\mathcal{L}'$. This outer loss measures the performance of the agent after the inner update(s), cross-validated on new samples. 
In this work, we consider looking ahead $n$ future inner updates from $\theta_t$ to $\theta_t^{(n)}$ to compute the meta-gradient $\bm{\nabla_{\eta_t}^{(n)}}$ at time $t$. We then keep the first inner update to change $\theta_t$ into $\theta_{t+1}$, and discard the $n-1$ other inner updates. The goal of this procedure is to be able to assess the quality of different meta-gradient updates (e.g. multi-step) while keeping the inner update unchanged. The gradient computation is detailed in figure~\ref{fig:meta_gradient_diagram} and equation~\ref{eq:theta_t^(n)}. 
\begin{equation}
\label{eq:theta_t^(n)}
    \eta_{t+1} = \eta_t - \beta \underbrace{\nabla_{\eta_t}\mathcal{L}'(\theta_t^{(n)}, \eta', \tau_t^{(n)})}_{\bm{\nabla_{\eta_t}^{(n)}}}
\quad
\text{with} 
\quad \begin{cases}
\theta_t^{(0)} = \theta_t\\
\tau_t^{(j)} \sim \pi_{\theta_t^{(j)}}, \quad 0 \leq j \leq n \\
\theta_t^{(j+1)} = \theta_t^{(j)} - \alpha \nabla_{\theta_t^{(j)}} \mathcal{L} \left(\theta_t^{(j)}, \eta_t, \tau_t^{(j)}\right)\\
\end{cases}
\end{equation}

Estimating an $n$-step meta-gradient is prone to high variance, as it requires evaluating the performance of $\theta_t^{(n)}$ which is the result of a succession of $n$ stochastic inner updates.
To assess this variance and measure it, we define the meta-gradient variance $\textrm{Var} \left[\bm{\nabla_{\eta_t}}\right]$ as the vector of the variances of each meta-parameter considered independently in the $n$-step meta-gradient estimation over the whole trajectory of inner updates $\left\{\theta_t^{(i)}\right\}_{1 \leq i \leq n}$.

\subsection{Meta-Learning Oracle}

Different choices of self-tuning or online meta-learning techniques give rise to varied learning dynamics and can lead to disparate performance. In equation~\ref{eq:oracle}, we aim to define an oracle that would give the best algorithm with which the agent may learn most effectively. At time $t$ during training, we define the oracle as the meta-parameter trajectory $\eta_{t:T-1} \equiv \{\eta_t, \dots, \eta_{T-1}\}$ that maximises the performance of the parameters at the end of training $\theta_T$.

\begin{multline}
\label{eq:oracle}
\eta_{\: t:T-1}^\star = \arg\! \min_{\eta_{t:T-1}} \mathbb{E}_{\{ \theta_t^{(T-t)}, \tilde{\tau}_t^{(T-t)} \}} \left[ \mathcal{L}' \left(\theta_t^{(T-t)} (\eta_{t:T-1}), \eta', \tilde{\tau}_t^{(T-t)}\right) \right] \\
\text{With} \quad
\begin{cases}
\theta_t^{(0)} = \theta_t \\
\tau_t^{(j)} \sim \pi_{\theta_t^{(j)}}, \quad j \in [0, T-t]\\
\theta_t^{(j+1)} (\eta_{t:t+j}) = \theta_t^{(j)} (\eta_{t:t+j-1}) - \alpha \nabla_{\theta_t^{(j)}} \mathcal{L} \left(\theta_t^{(j)}, \eta_{t+j}, \tau_t^{(j)} \right) \\
\end{cases}
\end{multline}


This oracle quickly becomes intractable as it requires optimising an expected value in a continuous, non-convex, and very high-dimensional space. In the experiments, we wish to measure how close to the oracle an algorithm may perform. As the oracle is computationally too expensive to be practical, we need to approximate it. By considering simple environments and restricting $n$ to a large finite number, we can use the expectation of the $n$-step meta-gradient, noted $\mathbb{E}\left[\bm{\nabla_{\eta_t}^{(n)}}\right]$, as a proxy for the oracle. The meta-gradient bias is thus defined as the difference between the oracle proxy and the expected meta-gradient.
\begin{equation}
    \textrm{Bias}\left[\bm{\nabla_{\eta_t}}\right] = \left\| \mathbb{E}\left[\bm{\nabla_{\eta_t}}\right] - \mathbb{E}\left[\bm{\nabla_{\eta_t}^{(n)}}\right] \right\|
\end{equation}


\section{Experiments \& Results}

In this work, we analyse how different hyper-parameter trajectories influence the learning of an agent. Therefore, we focus on different ways to estimate meta-gradients, yet we always keep the inner update fixed. This is to measure the effect on performance, of the quality of the estimation of meta-gradients.


\subsection{Environments}


\begin{figure}[h]
    \centering
    \begin{subfigure}[b]{0.55\textwidth}
        \centering
        \includegraphics[width=\textwidth]{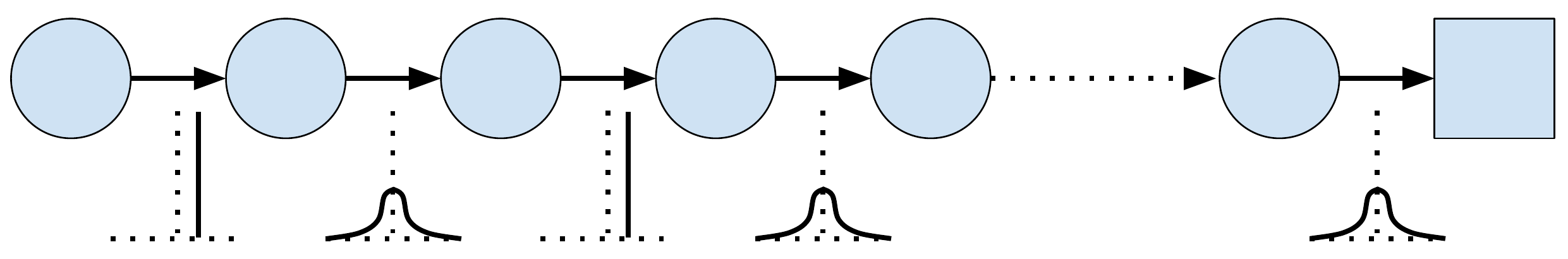}
        \caption{Markov Reward Process. Source: Xu et al. (2018)~\cite{xu2018metagradient}.}
        \label{fig:mrp_env}
    \end{subfigure}
    \hspace{3em}
    \begin{subfigure}[b]{0.21\textwidth}
        \centering
        \includegraphics[width=\textwidth]{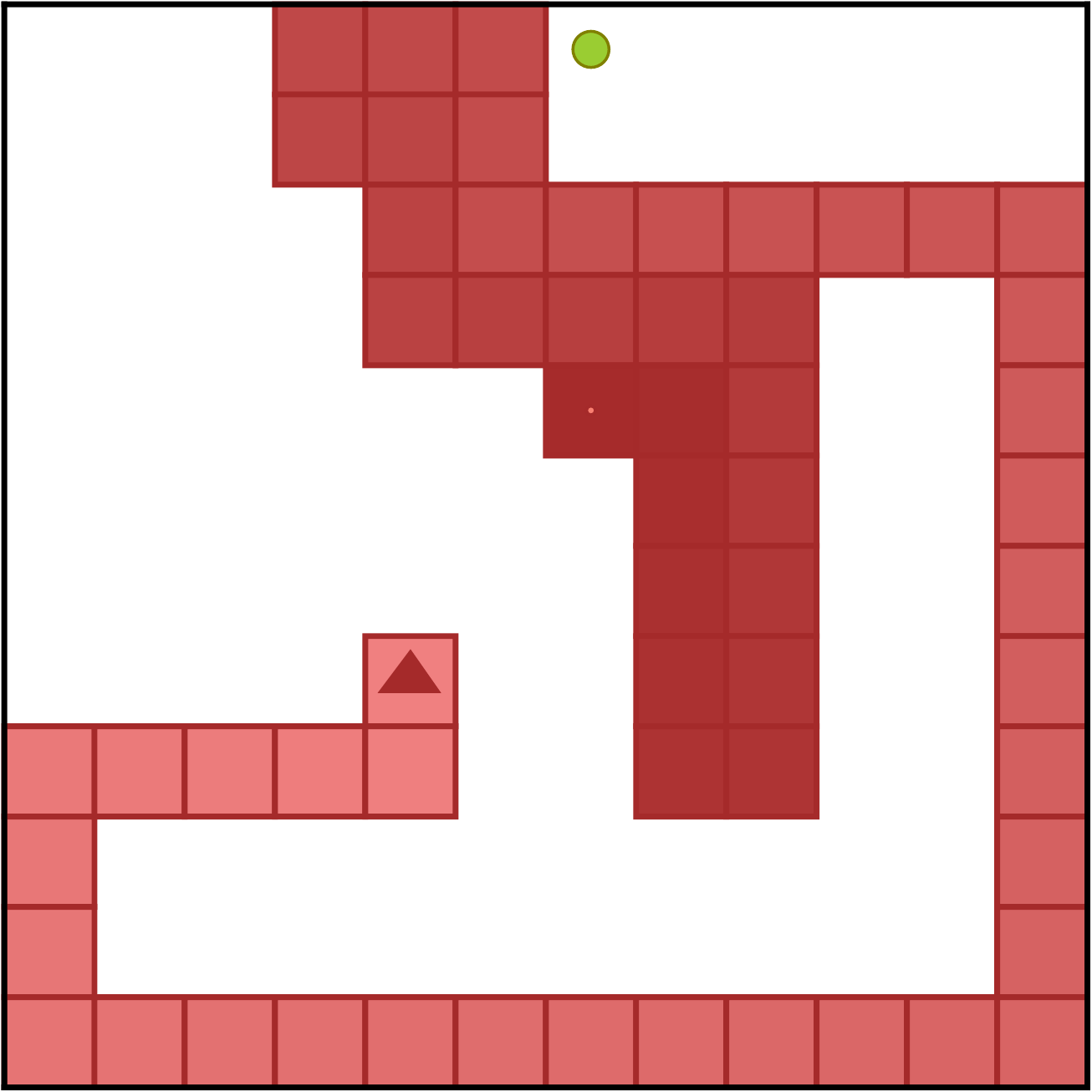}
        \caption{Snake Environment.}
        \label{fig:snake_env}
    \end{subfigure}
    \label{fig:envs}
    \caption{Environments used for the experiments. \textbf{(a)}: Markov Reward Process (MRP), with 10 states and transitions from left to right only. Even-numbered states output a deterministic reward of +0.1, while odd-numbered states provide a random reward sampled from the standard normal distribution. The goal of the agent is to predict the state-value function. A discount factor is meta-learnt for each state to enable the agent to discard the noisy rewards in the value estimation.
    \textbf{(b)}: Implementation of the Snake game on a 12x12 grid. The snake moves one cell at a time to collect fruits on the board. Its length grows by 1 with each collected fruit, making it critical for the agent to establish long-term strategies to survive.
    }
\end{figure}

For this purpose, we run experiments on two custom-built environments written in JAX~\cite{jax2018github}, whose detailed descriptions can be found in appendix~\ref{sec:env_appendix}. The first one is a Markov Reward Process (MRP) introduced in Xu et al. (2018)~\cite{xu2018metagradient} in which even-numbered states provide deterministic rewards and odd-numbered ones output rewards that are sampled from a zero-mean Gaussian (figure~\ref{fig:mrp_env}). Per-state discount factors $\gamma_i$
can be tuned to improve
the signal-to-noise ratio of the gradient of the TD($\lambda$) error by discarding noisy rewards in the computation of the state value function. Since no actions are taken, the MRP has a lower variance than typical RL problems, hence, it is well-suited for meta-learning experiments as meta-parameter trajectories are easier to predict. In this problem, effective meta-learning would see odd-numbered discount factors decrease to 0 whilst even-numbered ones increase to 1. The second environment implements the game of Snake (figure~\ref{fig:snake_env}). The latter contains a single agent, namely the snake, whose goal is to navigate in a 12x12 grid world to collect as many fruits as possible, without colliding with its own body, i.e. looping on itself. Its length grows by 1 with each fruit it gathers, making it harder to survive as the episode progresses. An episode ends if the snake exits the board, hits itself, or after 5000 steps.
Planning within an episode is directly linked to the discount factor. Hence, meta-learning $\gamma$ enables the Snake agent to start off with short planning horizons and progressively increase them as performance rises.

\subsection{Multi-Step Meta-Gradients}

\begin{table}[t]
    \centering
    {\renewcommand{\arraystretch}{1.2}%
    \centerline{
    \begin{tabular}{|c||c|c|c|c|c||c|c|}
        \hline
        Inner Steps $n$ & 0 & 1 & 3 & 5 & 10 & 3-mix & 5-mix \\
        \hline
        \hline
        Final Return & $67\;(3)$ & $85\;(28)$ & $\textbf{114}\;(15)$ & $97\;(7)$ & $86\;(9)$ & $\textbf{115}\;(2)$ & $113\;(12)$ \\
        \hline
        $\left\| \textrm{Std}\left[\bm{\nabla_{\eta}}\right] \right\|$ & --- & $1.0$ & $4.6\;(0.5)$ & $8.6\;(2.1)$ & $23.5\;(8.4)$ & $1.7\;(0.4)$ & $2.6\;(0.8)$ \\
        \hline
    \end{tabular}
    }
}
    \caption{Performance of $n$-step meta-gradients with varying $n$ on the \textbf{Snake} environment. Metrics are averaged over 3 different seeds with their means and standard deviations displayed in the table. The first row shows the number of look-ahead inner steps used for each iteration to compute the meta-gradient. $0$ inner steps refers to the baseline where the meta-parameters are fixed and do not evolve with time. The second row shows the return at the end of the training, with its deviation on 3 seeds. Finally, the last row gives the meta-gradient standard deviation normalised with respect to the 1-step update and averaged over the 4 meta-parameters $\{\gamma, \eta, c_{\textrm{crit}}, c_{\textrm{entr}}\}$. It means that, for instance, across all 4 meta-parameters, the meta-gradient deviation is on average 5 times higher for $10$-step meta-gradient than for $3$-step. The variance was periodically estimated using batches of 96 meta-gradients at 300 different points throughout training. The $n$-step mix algorithm is detailed in section~\ref{sec:mix_algorithm}.}
    \label{tab:snake_n_steps}
\end{table}

\begin{figure}[t!]
     \centering
     \begin{subfigure}[b]{0.61\textwidth}
        \centering
        \includegraphics[width=\textwidth]{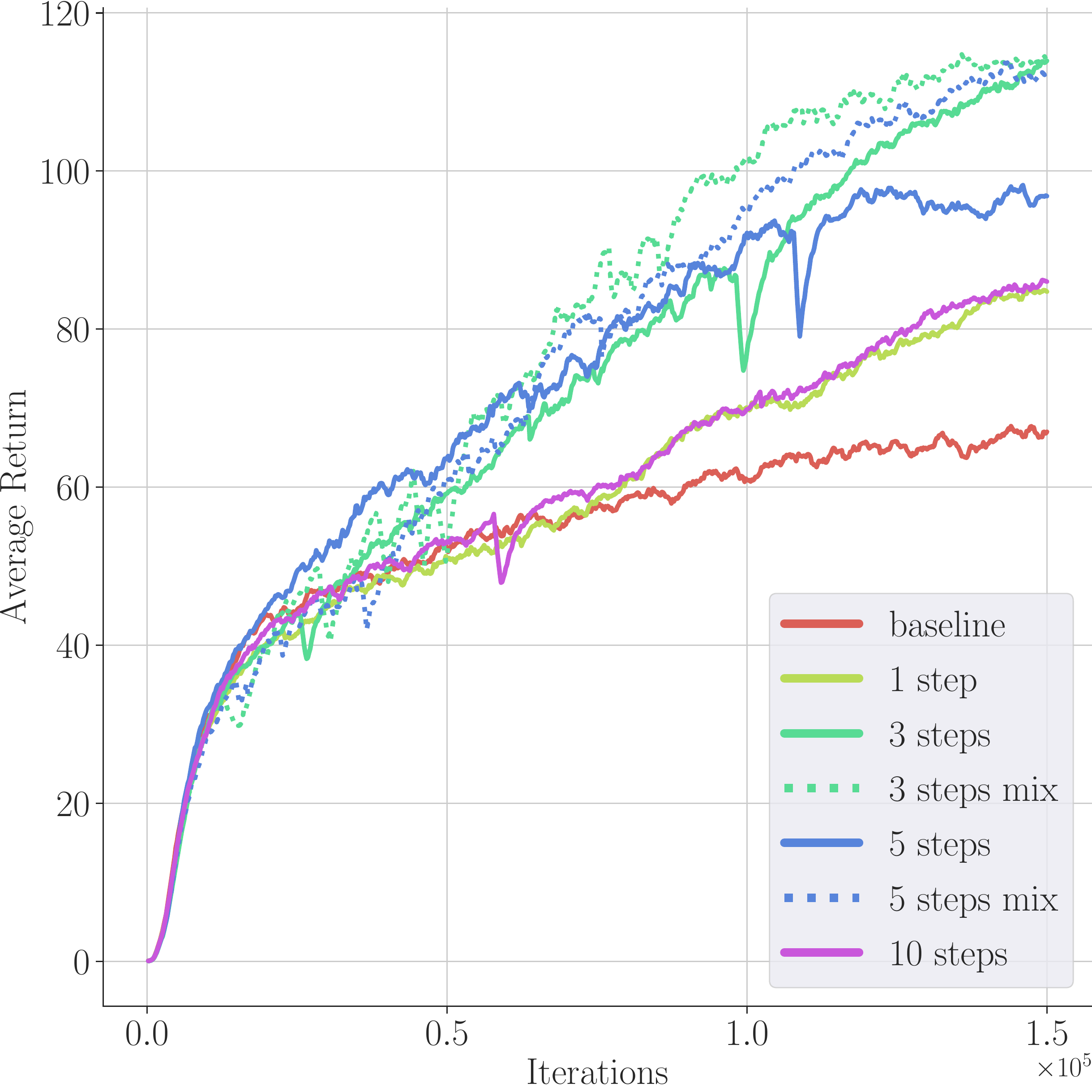}
        \caption{Average return}
        \label{fig:snake_n_steps_mix_return}
     \end{subfigure}
     \hfill
     \begin{subfigure}[b]{0.36\textwidth}
        \centering
        \includegraphics[width=\textwidth]{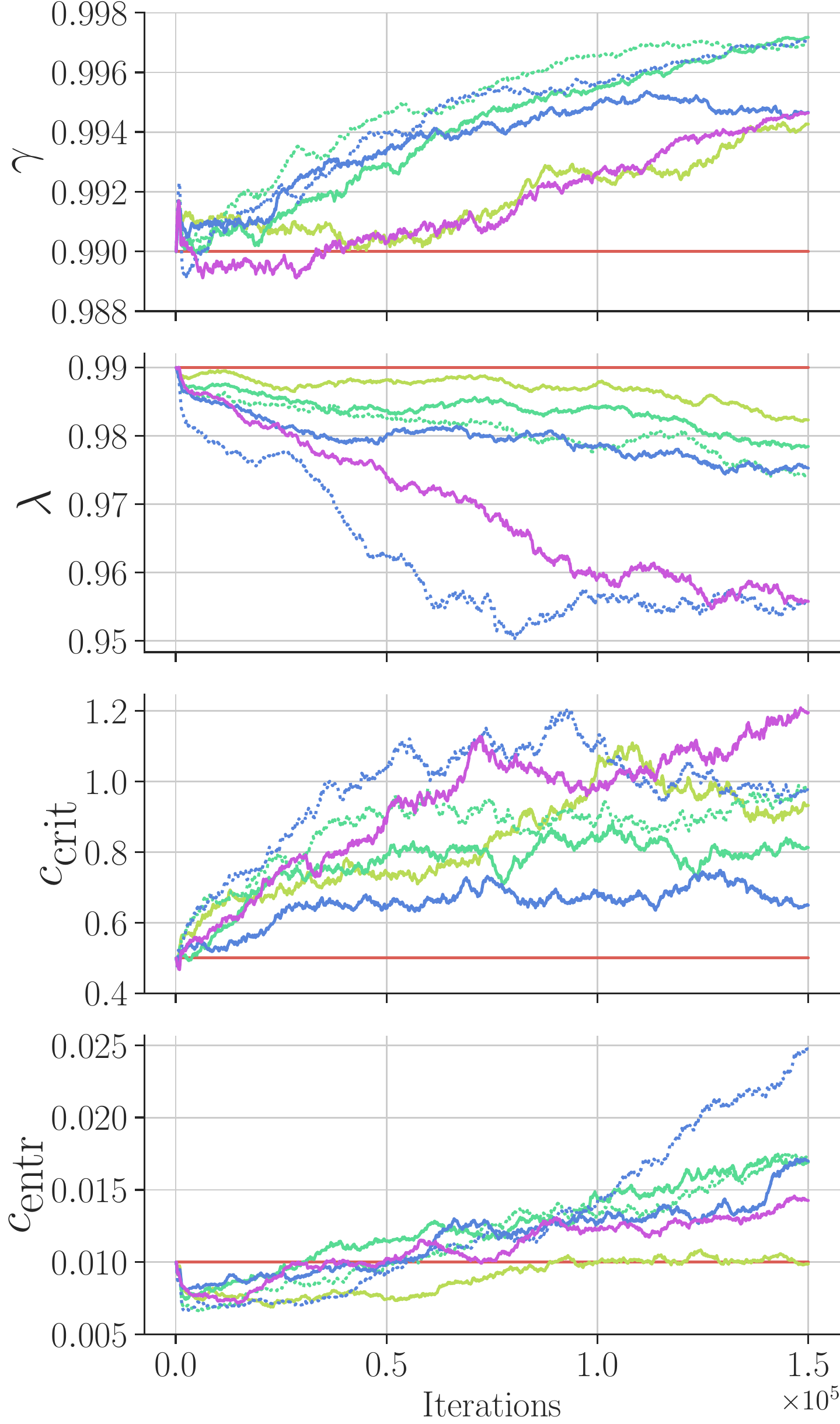}
        \caption{Meta-parameter trajectories}
        \label{fig:snake_n_steps_mix_meta_params}
     \end{subfigure}
    \caption{Learning curves for the \textbf{Snake} environment for different values $n$ of inner steps used for meta-gradients, lines represent the average over 3 seeds. Baseline means meta-parameters remain fixed during training. \textbf{(a)}: Average return throughout training. The theoretical maximum return is 143, when all fruits are collected on time. \textbf{(b)}: Learning curves of the meta-parameters.
    }
    \label{fig:n_steps_mix_plots}
\end{figure}

In table~\ref{tab:snake_n_steps}, we compare the difference in performance and meta-gradient variance when using different values of $n$ to compute $n$-step meta-gradients in the Snake environment. We observe that performance increases with $n$ to a point and then deteriorates, whereas meta-gradient variance keeps increasing with $n$, showing a strong correlation between the variance and the number of inner updates. Learning curves are displayed in figure~\ref{fig:n_steps_mix_plots} and show that meta-learning the discount factor $\gamma$ seems to yield the highest correlation with performance. When $n$ is too small, the meta-objective seems myopic, whereas when $n$ is too high, the trajectory appears noisier, in both cases slowing down the meta-learning of $\gamma$ and hence, hindering performance. The "mix" version is detailed later in section~\ref{sec:mix_algorithm}. In the two next sections, we analyse both the bias and the variance of $n$-step meta-gradients.

\subsection{Monte Carlo Meta-Gradient}

During each step of training, the agent first takes one or several gradient steps using a few rollouts, and then the meta-learning algorithm validates its performance on new rollouts to compute the meta-gradient. Its computation combines randomness from both the inner and outer loops, and therefore the meta-gradient has high variance, which may affect performance (as seen in table~\ref{tab:snake_n_steps}).
Here, we analyse the expected direction of the meta-gradient by averaging a batch of computed meta-gradients, i.e.
we calculate a Monte Carlo approximation of the expected meta-gradient.
In this way, we can compare different ways of computing the meta-gradient without yet taking the variance into account. For all algorithms, we keep the inner update of the network parameters the same, meaning we do not use more data to train the agent itself, but only to update the meta-parameters of the learning algorithms. The $n$-step Monte Carlo meta-update is given in equation~\ref{eq:mc_n_steps} with $\theta_t^{(n)}$ derived from equation~\ref{eq:theta_t^(n)}.

\begin{equation}
    \theta_{t+1} = \theta_t - \alpha \nabla_{\theta_t} \mathcal{L} (\theta_t, \eta_t, \tau_t), \qquad \tau_t \sim \pi_{\theta_t}
\end{equation}
\begin{equation}
\label{eq:mc_n_steps}
    \eta_{t+1} = \eta_t - \beta \nabla_{\eta_t} \mathbb{E}_{\{\tau_t^{(j)} \sim \pi_{\theta_t^{(j)}}\}_{j \in [0,n]}} \left[ \mathcal{L}'(\theta_t^{(n)}, \eta', \tau_t^{(n)}) \right]
\end{equation}

\begin{figure}[h]
     \centering
     \begin{subfigure}[b]{0.32\textwidth}
        \centering
        \includegraphics[width=\textwidth]{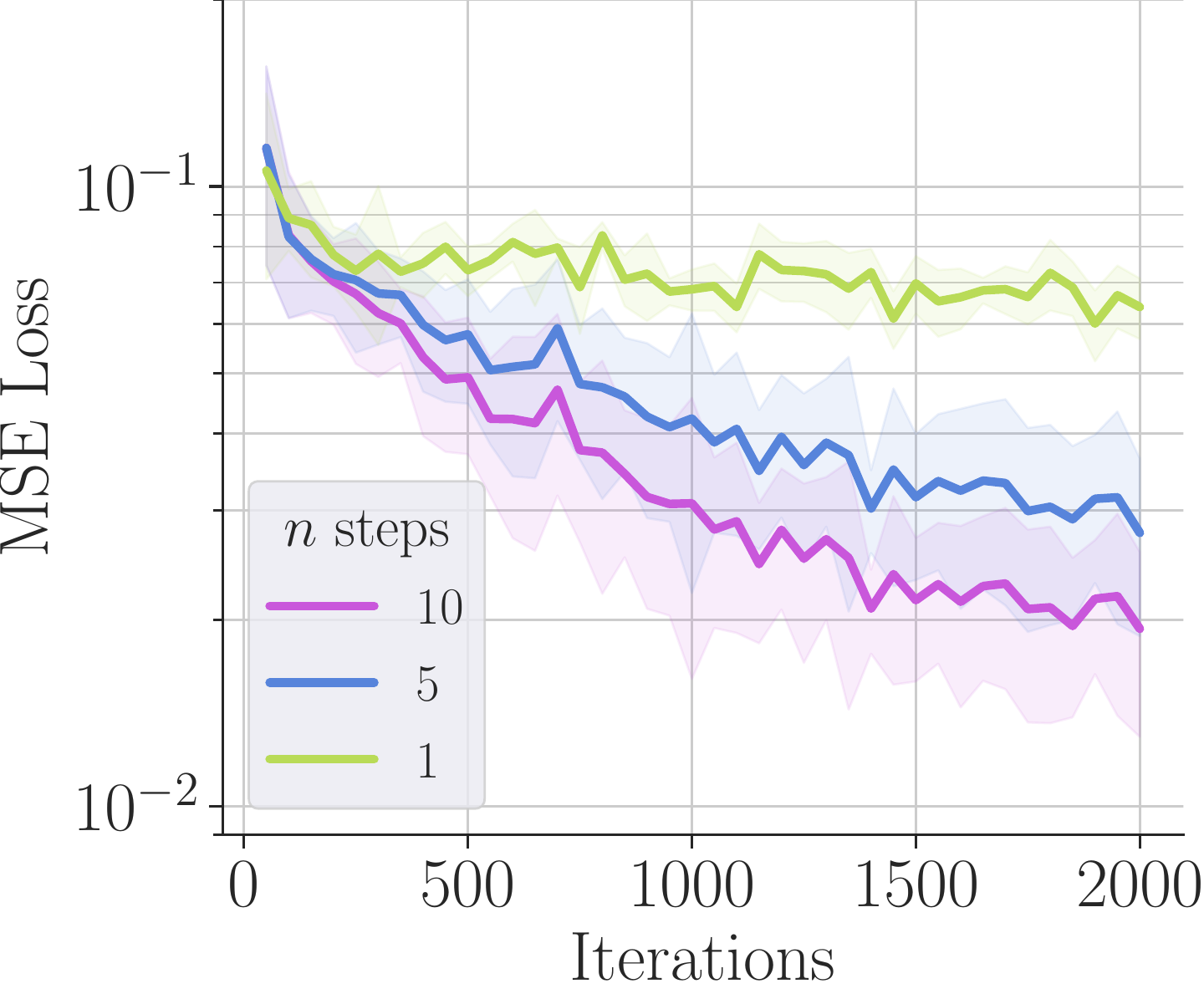}
        \caption{Training loss}
        \label{fig:mrp_mc_loss}
     \end{subfigure}
     \hfill
     \begin{subfigure}[b]{0.32\textwidth}
        \centering
        \includegraphics[width=\textwidth]{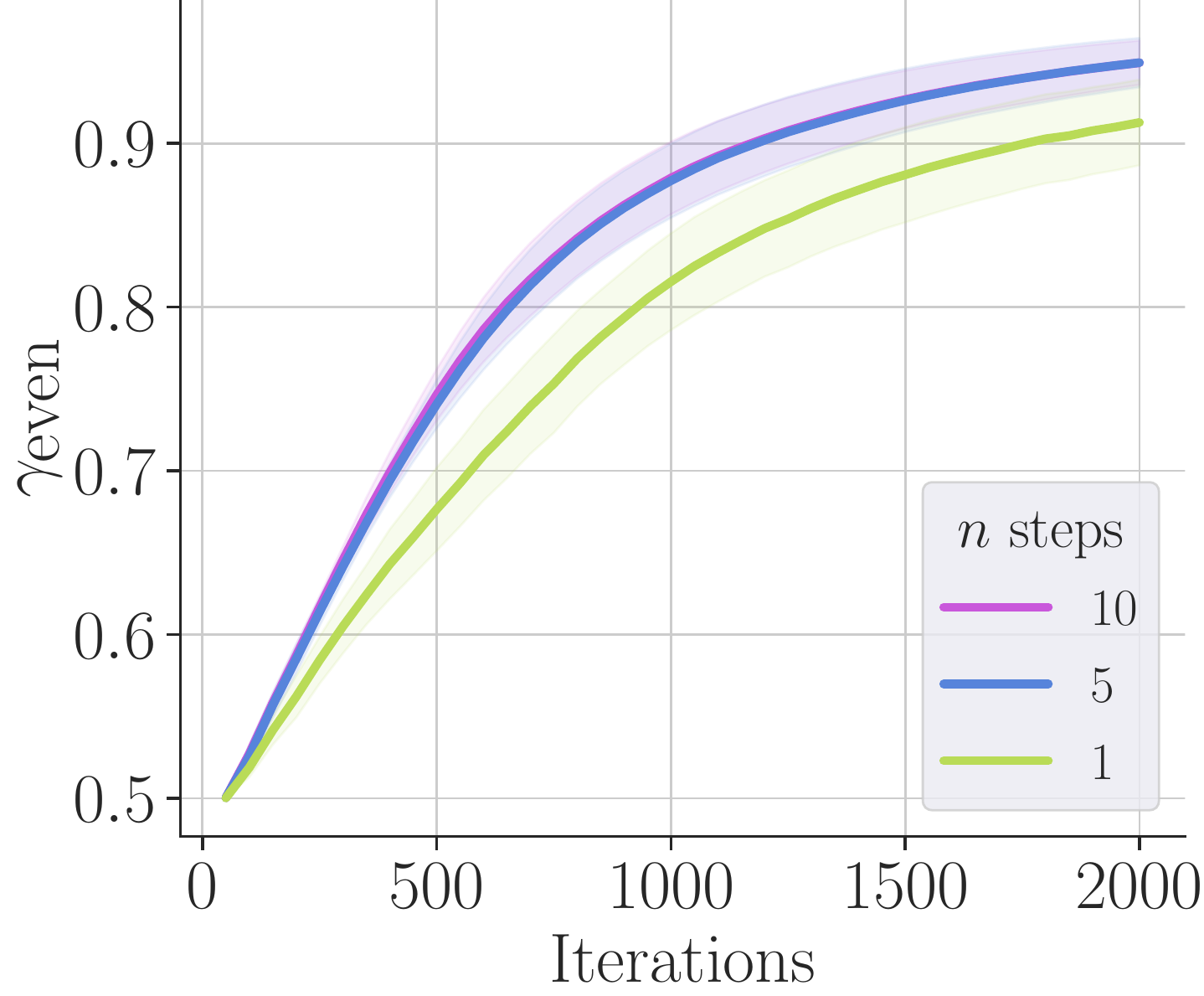}
        \caption{$\gamma_{\textrm{even}}$ Trajectories}
        \label{fig:mrp_mc_even_gammas}
     \end{subfigure}
     \hfill
     \begin{subfigure}[b]{0.32\textwidth}
        \centering
        \includegraphics[width=\textwidth]{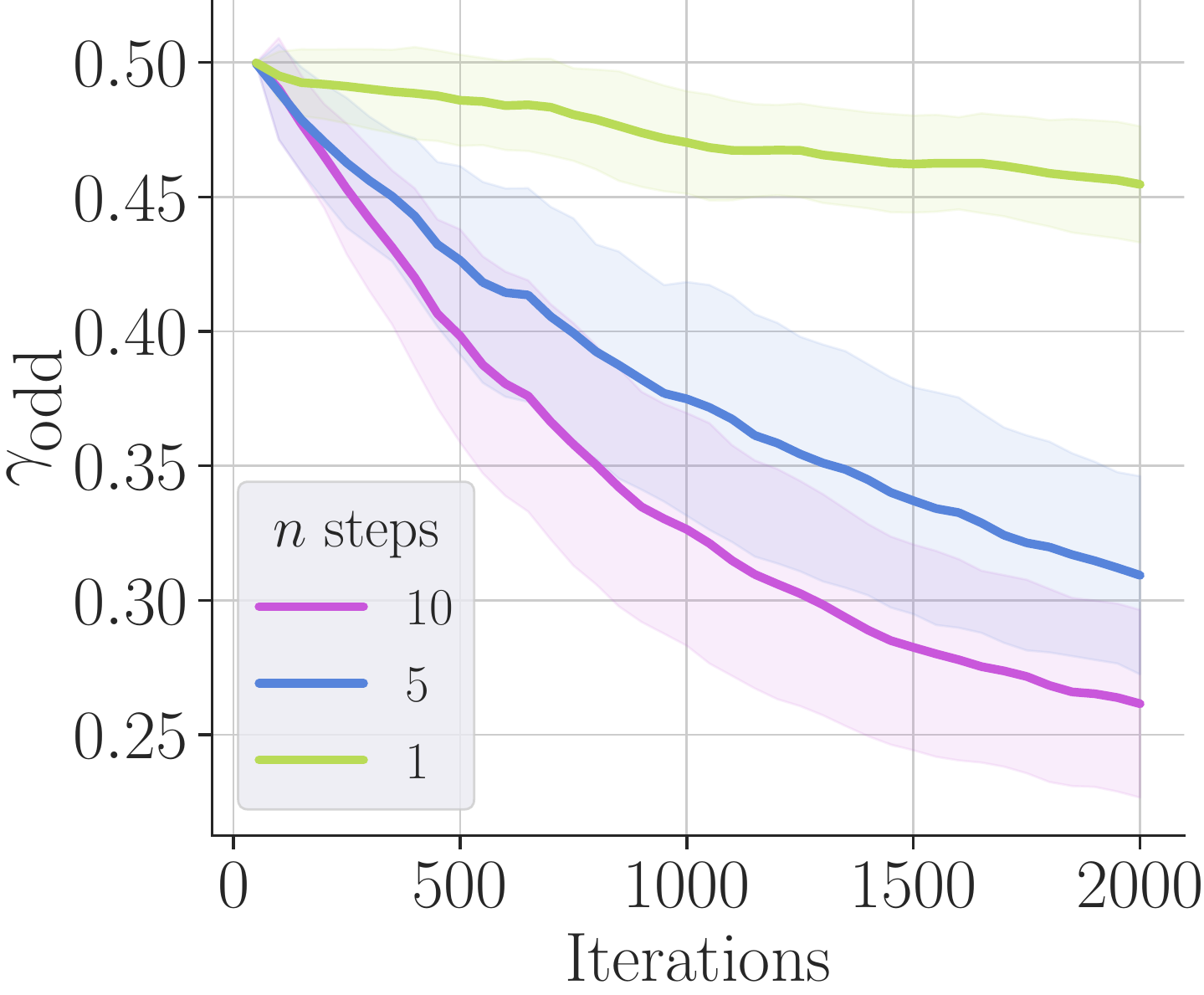}
        \caption{$\gamma_{\textrm{odd}}$ Trajectories}
        \label{fig:mrp_mc_odd_gammas}
     \end{subfigure}
        \caption{Markov Reward Process Monte Carlo meta-learning of per-state discount factors $\{\gamma_i\}_{i\in[0,9]}$. \textbf{(a)}: MSE training loss (log scale). \textbf{(b)}: Trajectories of even-numbered $\gamma$, supposed to increase to 1. The $5$- and $10$-step lines overlap as they are identical.  \textbf{(c)}: Trajectories of odd-numbered $\gamma$, meant to decrease to 0. At each iteration a Monte Carlo estimation of 256 meta-gradients is used for the meta-update, this is to ensure meta-gradient variance does not interfere in these experiments. Shaded areas represent the standard deviation over 5 random seeds. We observe that, taking the meta-gradient variance out, increasing $n$ leads to faster convergence and better performance.
        }
        \label{fig:mrp_mc_plots}
\end{figure}

We compare Monte Carlo meta-gradient estimations using $n \in \{1, 5, 10\}$ inner updates for the computation of each meta-gradient. To analyse
the expected meta-gradient signals,
we use a very high number of Monte Carlo simulations, i.e. 256, at every step of training, which can only be done in simple toy problems such as the introduced Markov Reward Process.
Returns and meta-parameter trajectories are shown in figure~\ref{fig:mrp_mc_plots} where one can already see evidence that the expected meta-gradients are more informative, as they result in higher performance when using more inner updates.

\subsection{Bias and Variance of Meta-Gradients}
\label{sec:bias_variance}

We have seen that taking more training steps into account while computing meta-gradients may lead to higher variance which can hinder performance (table~\ref{tab:snake_n_steps}), yet it may also provide a better signal for the learning of the agent (figure~\ref{fig:mrp_mc_plots}).

\begin{figure}
     \centering
     \begin{subfigure}[b]{0.32\textwidth}
        \centering
        \includegraphics[width=\textwidth]{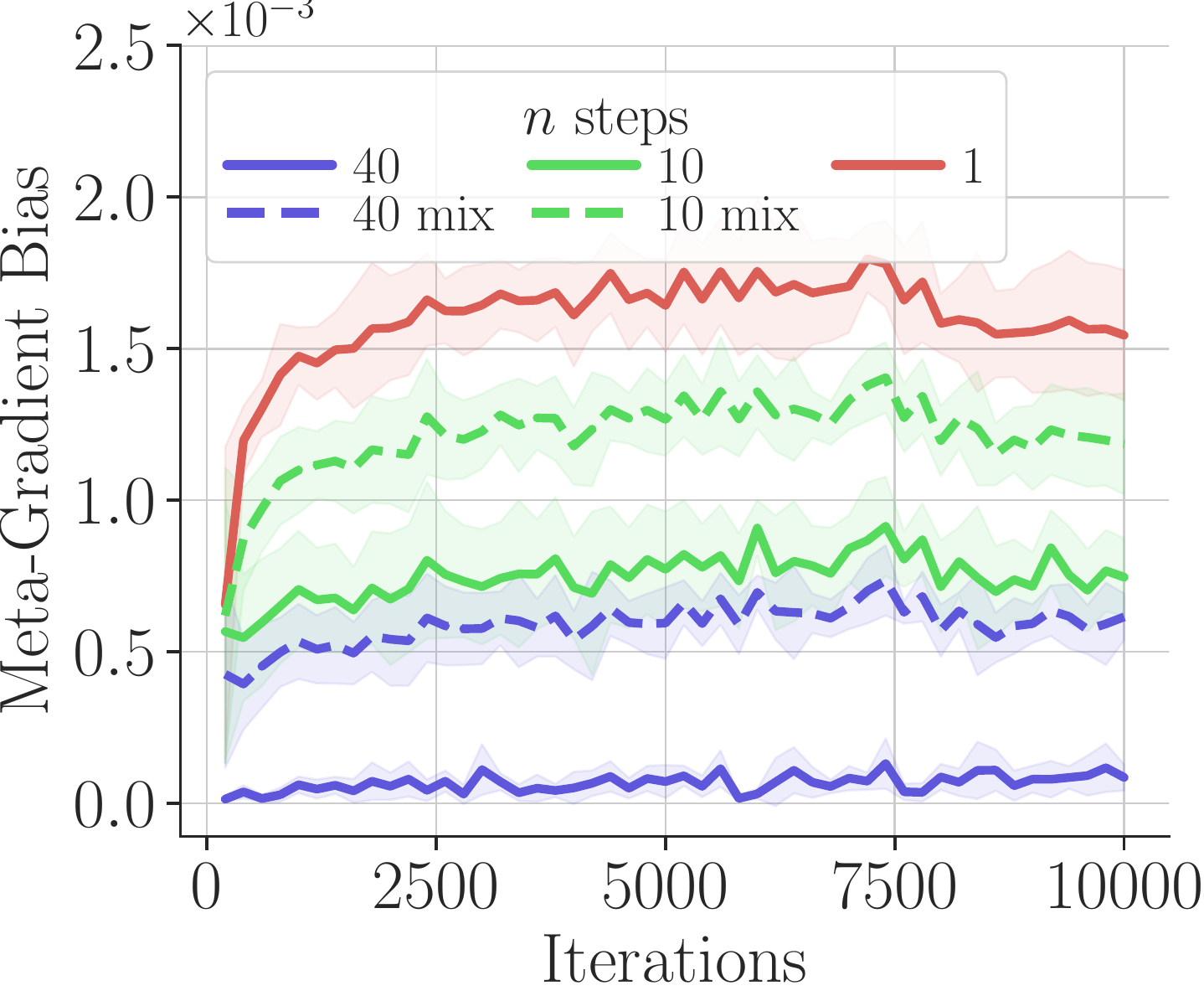}
        \caption{Meta-Gradient Bias}
        \label{fig:meta_gradient_bias}
     \end{subfigure}
     \hfill
     \begin{subfigure}[b]{0.32\textwidth}
        \centering
        \includegraphics[width=\textwidth]{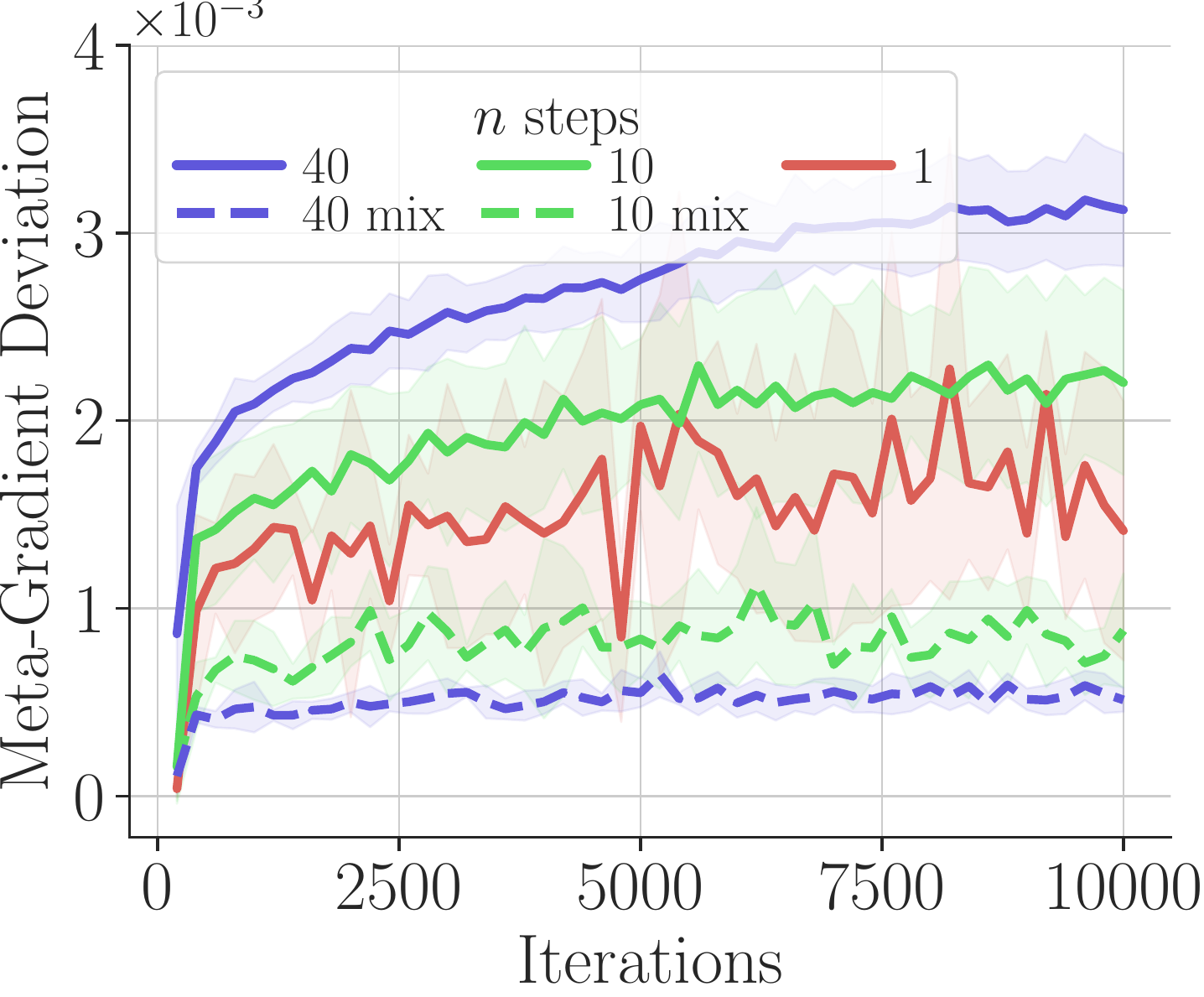}
        \caption{Meta-Gradient Deviation}
        \label{fig:meta_gradient_variance}
     \end{subfigure}
     \hfill
     \begin{subfigure}[b]{0.32\textwidth}
        \centering
        \includegraphics[width=\textwidth]{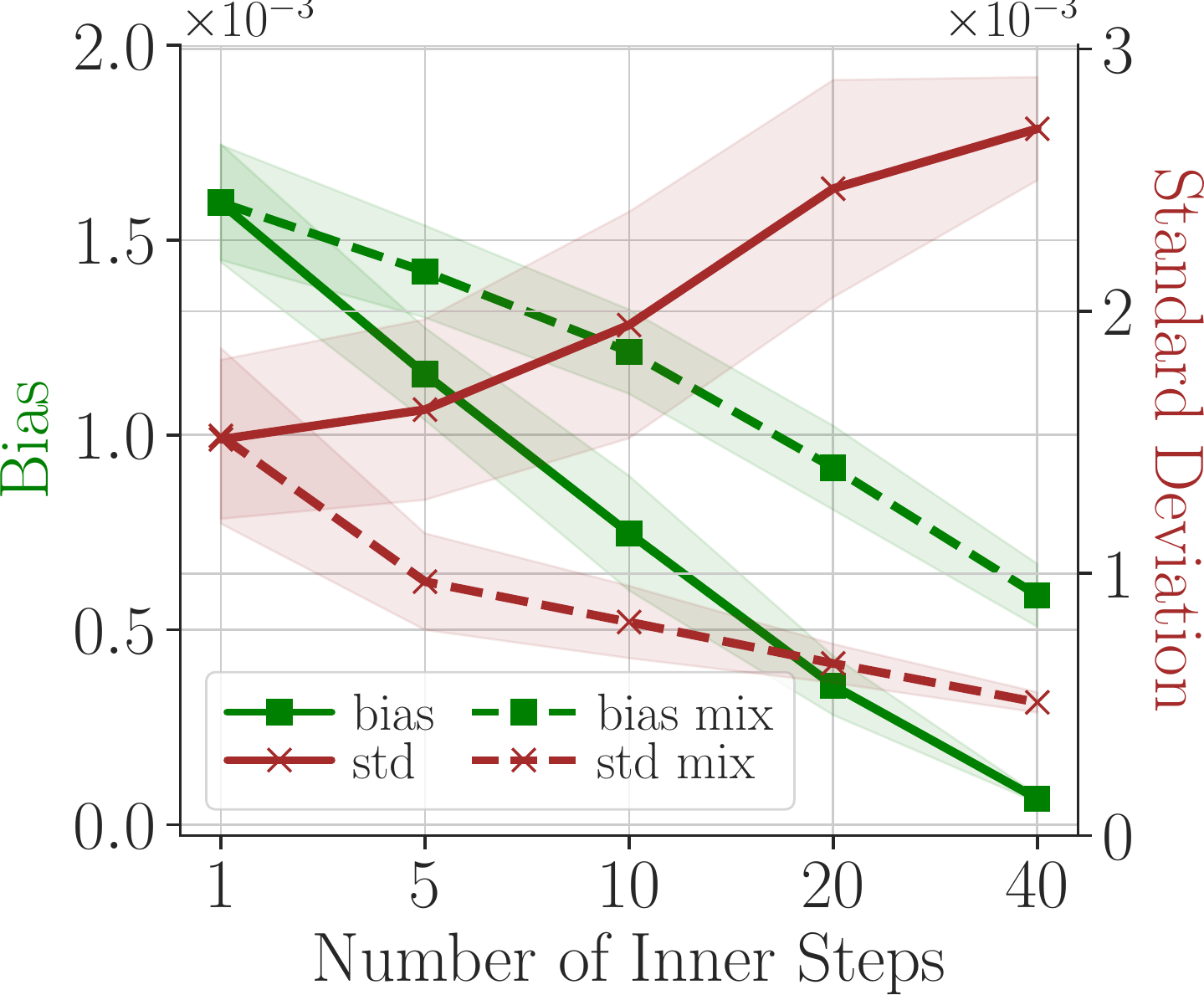}
        \caption{Bias - Variance}
        \label{fig:bias_variance_plot}
     \end{subfigure}
        \caption{Markov Reward Process bias-variance experiment. Along the training trajectory of a standard $1$-step meta-gradient algorithm, Monte Carlo meta-gradient updates are plotted for different values of $n$. Shaded areas represent the standard deviation over 5 random seeds, averaged over all even-numbered $\gamma_i$. Each Monte Carlo estimation uses 2048 meta-gradients to precisely estimate the bias and variance. \textbf{(a)}: Meta-gradient bias throughout training, using $40$-step Monte Carlo meta-gradient as the oracle. \textbf{(b)}: Meta-gradient standard deviation throughout training. \textbf{(c)}: Both deviation and bias averaged over the whole training and plotted as a function of $n$ for both the classic $n$-step meta-gradient and mixed algorithms, the latter being explained in section~\ref{sec:mix_algorithm}.
        }
        \label{fig:bias_variance_plots}
\end{figure}

We further analyse the meta-gradients themselves by inspecting their bias and variance throughout training in figure~\ref{fig:bias_variance_plots}. To do so, we compute what the Monte Carlo meta-gradient updates would be for different values of $n$ along the training trajectory of a standard $1$-step meta-gradient algorithm.

Although computing the variance from a batch of meta-gradient estimates is rather straightforward, particular attention must be given to the estimation of the bias. Unfortunately, the oracle described in equation~\ref{eq:oracle} is intractable. Therefore, we are forced to bootstrap it and estimate the bias by looking at $n$ steps of training instead. Consequently, the meta-gradient used as an oracle in the experiments is the $n$-step Monte Carlo expression (with $n=40$) given in equation~\ref{eq:mc_n_steps}.

In the bias-variance experiments, one may see that the higher the $n$, the higher the variance, yet the lower the bias, providing a better expected meta-gradient in the sense of closer to the oracle. Therefore, we highlight a bias-variance trade-off analogous to $n$-step TD learning~\cite{sutton2018reinforcement}.

\subsection{Trading Off Bias and Variance}
\label{sec:mix_algorithm}

Similarly to TD($\lambda$)~\cite{sutton2018reinforcement}, we propose to control the meta-gradient bias-variance trade-off by mixing all $i$-step updates with a discount factor $\kappa \in [0,1]$. This mixing version gives more weight to $i$-step meta-gradient estimates with lower $i$. The mixing update computes all $i$-step meta-gradients and discounts them by $\kappa$ such that their weight ranges from $\kappa^0$ to $\kappa^{n-1}$. A normalising constant $\frac{1-\kappa}{1-\kappa^n}$ is included so that the mixing meta-gradient norm does not explode with $n$. In equation~\ref{eq:n_steps_mix}, the $i^{\textrm{th}}$ inner update $\theta_t^{(i)}$ is derived from equation~\ref{eq:theta_t^(n)}.
\begin{equation}
\label{eq:n_steps_mix}
    \eta_{t+1} = \eta_t - \beta \frac{1-\kappa}{1-\kappa^n} \sum_{i=1}^n \kappa^{i-1} \nabla_{\eta_t} \mathcal{L}'(\theta_t^{(i)}, \eta', \tilde{\tau}_t^{(i)})
\end{equation}

Although this mixing $n$-step algorithm brings a new hyper-parameter $\kappa$ that controls the horizon of inner updates through which meta-gradients flow, we can automatically set it such that $(1 - \kappa)^{-1} = n $
, similar to the choice of discount factor $\gamma$ as a function of the horizon in RL environments.
$\kappa$ could be fine-tuned over a grid-search, yet we find that this automatic baseline value performs well. For $\kappa = 0$, one retrieves the $1$-step meta-gradient algorithm with high bias and low variance, whereas $\kappa=1$ gives the same weight to all $i$-step meta-gradients. We compare this new algorithm with classic $n$-step updates in table~\ref{tab:snake_n_steps} and figure~\ref{fig:n_steps_mix_plots}.

In the Snake experiments, $n$-step meta-gradients seem to perform the best with $n = 3$
Indeed, the meta-gradient variance explodes with higher numbers of inner steps, making $5$-step and $10$-step underperform.
Introducing the mixing algorithm for the $n=3$ appears to perform similarly while cutting the variance by 3. Additionally, using the mixing with $n=5$ reduces the variance by the same amount, hence performing better than classic $5$-step meta-gradients. Therefore, mixing $n$-step meta-gradients appears as a way to control meta-gradient variance while maintaining a low bias, leading to robust and high performance.


\section{Discussion}

In the experiments, we could see the potential of multiple-step meta-gradients, yet we showed why they are not used in practice due to their high variance.

\subsection{Hessian Matrix}

To better analyse the origin of this variance, we decompose the gradient of the $n$-step outer loss $\mathcal{L}'$, whose derivation can be found in the appendix (\ref{sec:n_step_meta_gradient_formula}).

\begin{equation}
\label{eq:meta_gradient_hessian}
\nabla_{\eta_t} \mathcal{L}' (\theta_{t+n}, \eta', \tilde{\tau}_{t+n}) = T_{t+n} \sum_{i=1}^{n} \left( \prod_{j=1}^{i-1} \left(I + H_{t+n-j}\right) \right) N_{t+n-i}
\end{equation}
\begin{equation*}
\text{With} \begin{cases}
T_{t+n} = \frac{\partial \mathcal{L}'}{\partial \theta_{t+n}} (\theta_{t+n}, \eta', \tilde{\tau}_{t+n}) \\
H_{t+n-j} = - \alpha \frac{\partial^2 \mathcal{L} }{\partial \theta_{t+n-j}^2}(\theta_{t+n-j}, \eta_t, \tau_{t+n-j}) \\
N_{t+n-i} = - \alpha \frac{\partial^2 \mathcal{L} }{\partial \theta_{t+n-i} \partial \eta_t}(\theta_{t+n-i}, \eta_t, \tau_{t+n-i}) \\
\end{cases}
\end{equation*}

If we call $H_{t+n-j}$ the Hessian as it is the Hessian matrix, up to a factor $-\alpha$, of the inner loss $\mathcal{L}$ with respect to the parameters $\theta_{t+n-j}$ of the network, we show in the appendix (\ref{sec:hessian_and_meta_gradient_variance}) that products of such Hessian matrices may be responsible for the meta-gradient variance increase. Indeed, spectral radii may explode with the number of Hessian to multiply together. Further work could focus on these matrices $H_{t+n-j}$ to derive properties that would clearly explain the variance. One may also use the accumulative trace approximation~\cite{xu2018metagradient} to analyse the impact of the Hessian matrices on the meta-gradient variance and derive better methods for meta-gradients.

\subsection{Sample Efficiency}

The experiments were designed to keep the inner learning update unchanged but to work on the learning signal to provide to the agent. Therefore, the ideas presented here are not particularly sample efficient as they need to look ahead in the training dynamics of the agent to eventually provide it with a learning signal. However, we believe these ideas apply to most meta-gradient methods and future work could derive sample-efficient approaches to compute meta-gradients.
Consistent with our findings, recent work~\cite{flennerhag2021bootstrapped} managed to look ahead a few inner steps and then reuse the collected data to improve upon myopic meta-objectives. We would be excited to see efforts that use past inner updates to compute meta-gradients by looking backwards instead of forward in time, in either an off-policy or close to on-policy fashion.


\section{Conclusion}

Meta-gradients offer an elegant way of self-tuning a learning algorithm. Up to now, they have rarely been used with more than one inner update. In this work, we show that increasing the number of inner steps to differentiate through leads to higher meta-gradient variance, which may deteriorate its signal-to-noise ratio and hinder performance.
Yet, when using a close approximation of expected meta-gradients,
we observe the opposite trend in which performance is increased with a higher number of inner updates.
This means that, while a meta-gradient meta-learning algorithm is more adept to understand the training dynamics from multiple gradient steps rather than from only one, it may also suffer from higher variance.

As a result, we show the potential of trading off meta-gradient bias and variance to self-tune reinforcement learning agents within a single lifetime. We propose to mix multiple $n$-step meta-gradients to achieve lower variance while maintaining high performance, leading to more robust learning. We believe that better meta-learning algorithms may be designed if both meta-gradient bias and variance are understood and properly studied.


\nocite{*}
\bibliographystyle{plain}
\bibliography{references}

\begin{thebibliography}{10}

\bibitem{amit2020discount}
Ron Amit, Ron Meir, and Kamil Ciosek.
\newblock Discount factor as a regularizer in reinforcement learning, 2020.

\bibitem{bechtle2021metalearning}
Sarah Bechtle, Artem Molchanov, Yevgen Chebotar, Edward Grefenstette, Ludovic
  Righetti, Gaurav Sukhatme, and Franziska Meier.
\newblock Meta-learning via learned loss, 2021.

\bibitem{BertsekasTsitsiklis96}
D.~P. Bertsekas and J.~N. Tsitsiklis.
\newblock {\em Neuro-dynamic programming.}
\newblock Athena Scientific, 1996.

\bibitem{jax2018github}
James Bradbury, Roy Frostig, Peter Hawkins, Matthew~James Johnson, Chris Leary,
  Dougal Maclaurin, George Necula, Adam Paszke, Jake Vander{P}las, Skye
  Wanderman-{M}ilne, and Qiao Zhang.
\newblock {JAX}: composable transformations of {P}ython+{N}um{P}y programs,
  2018.

\bibitem{douglas_pearcy_1970}
R.~G. Douglas and Carl Pearcy.
\newblock On the spectral theorem for normal operators.
\newblock {\em Mathematical Proceedings of the Cambridge Philosophical
  Society}, 68(2):393–400, 1970.

\bibitem{finn2017modelagnostic}
Chelsea Finn, Pieter Abbeel, and Sergey Levine.
\newblock Model-agnostic meta-learning for fast adaptation of deep networks,
  2017.

\bibitem{flennerhag2021bootstrapped}
Sebastian Flennerhag, Yannick Schroecker, Tom Zahavy, Hado van Hasselt, David
  Silver, and Satinder Singh.
\newblock Bootstrapped meta-learning, 2021.

\bibitem{gupta2018metareinforcement}
Abhishek Gupta, Russell Mendonca, YuXuan Liu, Pieter Abbeel, and Sergey Levine.
\newblock Meta-reinforcement learning of structured exploration strategies,
  2018.

\bibitem{houthooft2018evolved}
Rein Houthooft, Richard~Y. Chen, Phillip Isola, Bradly~C. Stadie, Filip Wolski,
  Jonathan Ho, and Pieter Abbeel.
\newblock Evolved policy gradients, 2018.

\bibitem{kirsch2020improving}
Louis Kirsch, Sjoerd van Steenkiste, and Jürgen Schmidhuber.
\newblock Improving generalization in meta reinforcement learning using learned
  objectives, 2020.

\bibitem{Konda00actor-criticalgorithms}
Vijay Konda and John Tsitsiklis.
\newblock Actor-critic algorithms.
\newblock In {\em SIAM Journal on Control and Optimization}, pages 1008--1014.
  MIT Press, 2000.

\bibitem{metz2019understanding}
Luke Metz, Niru Maheswaranathan, Jeremy Nixon, C.~Daniel Freeman, and Jascha
  Sohl-Dickstein.
\newblock Understanding and correcting pathologies in the training of learned
  optimizers, 2019.

\bibitem{nichol2018firstorder}
Alex Nichol, Joshua Achiam, and John Schulman.
\newblock On first-order meta-learning algorithms, 2018.

\bibitem{oh2021discovering}
Junhyuk Oh, Matteo Hessel, Wojciech~M. Czarnecki, Zhongwen Xu, Hado van
  Hasselt, Satinder Singh, and David Silver.
\newblock Discovering reinforcement learning algorithms, 2021.

\bibitem{rakelly2019efficient}
Kate Rakelly, Aurick Zhou, Deirdre Quillen, Chelsea Finn, and Sergey Levine.
\newblock Efficient off-policy meta-reinforcement learning via probabilistic
  context variables, 2019.

\bibitem{schulman2018highdimensional}
John Schulman, Philipp Moritz, Sergey Levine, Michael Jordan, and Pieter
  Abbeel.
\newblock High-dimensional continuous control using generalized advantage
  estimation, 2018.

\bibitem{sutton2018reinforcement}
Richard~S Sutton and Andrew~G Barto.
\newblock {\em Reinforcement learning: An introduction}.
\newblock MIT press, 2018.

\bibitem{veeriah2019discovery}
Vivek Veeriah, Matteo Hessel, Zhongwen Xu, Richard Lewis, Janarthanan
  Rajendran, Junhyuk Oh, Hado van Hasselt, David Silver, and Satinder Singh.
\newblock Discovery of useful questions as auxiliary tasks, 2019.

\bibitem{wang2020exponentially}
Yufei Wang, Qiwei Ye, and Tie-Yan Liu.
\newblock Beyond exponentially discounted sum: Automatic learning of return
  function, 2020.

\bibitem{wu2018understanding}
Yuhuai Wu, Mengye Ren, Renjie Liao, and Roger Grosse.
\newblock Understanding short-horizon bias in stochastic meta-optimization,
  2018.

\bibitem{xu2020metagradient}
Zhongwen Xu, Hado van Hasselt, Matteo Hessel, Junhyuk Oh, Satinder Singh, and
  David Silver.
\newblock Meta-gradient reinforcement learning with an objective discovered
  online, 2020.

\bibitem{xu2018metagradient}
Zhongwen Xu, Hado van Hasselt, and David Silver.
\newblock Meta-gradient reinforcement learning, 2018.

\bibitem{zahavy2021selftuning}
Tom Zahavy, Zhongwen Xu, Vivek Veeriah, Matteo Hessel, Junhyuk Oh, Hado van
  Hasselt, David Silver, and Satinder Singh.
\newblock A self-tuning actor-critic algorithm, 2021.

\bibitem{zheng2018learning}
Zeyu Zheng, Junhyuk Oh, and Satinder Singh.
\newblock On learning intrinsic rewards for policy gradient methods, 2018.

\bibitem{zhou2020online}
Wei Zhou, Yiying Li, Yongxin Yang, Huaimin Wang, and Timothy~M. Hospedales.
\newblock Online meta-critic learning for off-policy actor-critic methods,
  2020.

\bibitem{zintgraf2020varibad}
Luisa Zintgraf, Kyriacos Shiarlis, Maximilian Igl, Sebastian Schulze, Yarin
  Gal, Katja Hofmann, and Shimon Whiteson.
\newblock Varibad: A very good method for bayes-adaptive deep rl via
  meta-learning, 2020.

\bibitem{zintgraf2019fast}
Luisa~M Zintgraf, Kyriacos Shiarlis, Vitaly Kurin, Katja Hofmann, and Shimon
  Whiteson.
\newblock Fast context adaptation via meta-learning, 2019.

\end{thebibliography}

\newpage
\appendix

\section{Reinforcement Learning Background}
For reinforcement learning, we consider the framework of a Markov Decision Process, which is a tuple $(S, A, P, R, \mu)$, where $S$ is the state space, $A$ the action space, $P$ the probability transition matrix, $R$ the reward function, and $\mu$ the initial state distribution. The goal of a reinforcement learning agent is to maximise the expected return which is the discounted sum of rewards.
\begin{equation*}
    \textrm{maximise} \; J = \mathbb{E}\left[\sum_{t=0}^{\infty} \gamma^t R_t\right]
\end{equation*}

In our analysis, we also integrate a simpler framework called Markov Reward Process in which there is no action space $A$. Hence, the agent cannot act in the environment but can only observe it.

\section{Environment Specifications}
\label{sec:env_appendix}
\subsection{Markov Reward Process}
The Markov Reward Process (MRP) used in the experiments is taken from Xu et al. (2018)~\cite{xu2018metagradient}. It consists of 10 states with transitions from left to right. Even-numbered states output a deterministic reward of $0.1$, whereas odd-numbered states give a random reward sampled from the standard normal distribution. Since the goal of the agent is to predict the state value function, the odd-numbered states make the task harder as they just add noise to the estimated targets.

\subsection{Snake}
\begin{figure}[h]
    \centering
    \begin{subfigure}[b]{0.14\textwidth}
        \centering
        \includegraphics[width=\textwidth]{figures/snake_env.pdf}
        \caption{Image}
    \end{subfigure}
    \hspace{3em}
    \begin{subfigure}[b]{0.75\textwidth}
        \centering
        \includegraphics[width=\textwidth]{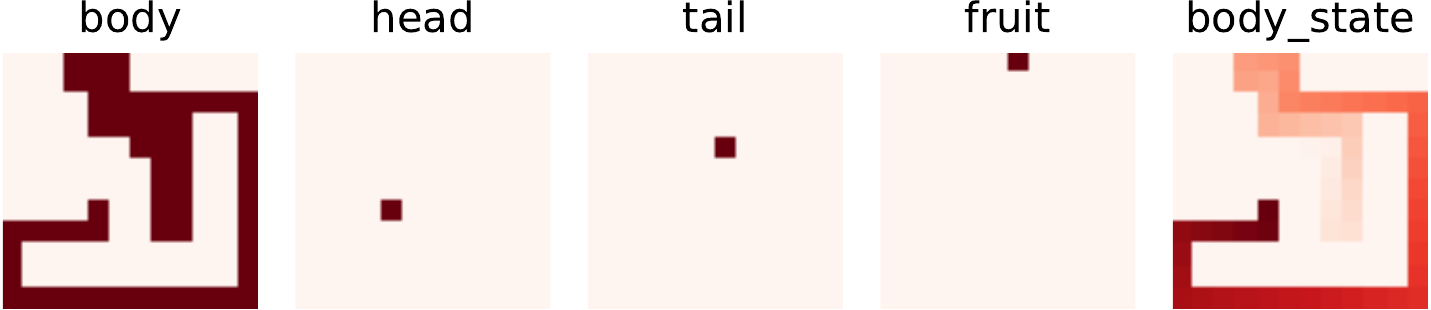}
        \caption{Observation}
        \label{fig:snake_obs}
    \end{subfigure}
    \caption{Snake environment.}
\end{figure}

The Snake environment consists of an agent, the snake, navigating a 2D grid of size $12\times12$. Its goal is to collect fruits on the board. The reward is 1 upon collection of each fruit, else 0. As an observation, the agent has access to the concatenation of 5 feature maps as channels stacked up in an image (figure~\ref{fig:snake_obs}) representing the snake body, its head, its tail, where the fruit is, as well as the order in which the cells are organised.

The environment is built using JAX~\cite{jax2018github} so that the training can be compiled on the device directly, allowing high computation speeds. The theoretical maximal reward is $12\times12 -1 = 143$ when the snake succeeds in collecting all fruits that have appeared on the grid.

\section{Experiments Details}

\subsection{Markov Reward Process}
For the Markov Reward Process experiments, we just use a value network for prediction. Both inner and outer losses are Mean Squared Error functions whose derivatives can be found below.

\begin{equation*}
    \frac{\partial \mathcal{L}(\theta, \eta, \tau)}{\partial \theta} =  - (g_\eta(\tau) - v_\theta(\tau)) \frac{\partial v_\theta(\tau)}{\partial \theta}
\end{equation*}

\begin{equation*}
    \frac{\partial \mathcal{L}' (\theta', \eta', \tau')}{\partial \eta} = - (g_{\eta'}(\tau) - v_{\theta'}(\tau)) \frac{\partial v_{\theta'}(\tau)}{\partial \eta}
\end{equation*}
For states $s_0$ to $s_9$, the targets $g_\eta(s_i)$ use per-state discount factors $\gamma_i$.
\begin{equation*}
    g_\eta(s_i) = \sum_{j=i}^{9} \gamma_j^{j-i+1} r_j
\end{equation*}
Whereas the outer target is just the Monte Carlo return, i.e. the undiscounted sum of rewards.
\begin{equation*}
    g_{\eta'}(s_i) = \sum_{j=i}^{9} r_j
\end{equation*}

Since some discount factors would converge to 0 and others to 1, we initialise all factors to 0.5 to observe both dynamics.

\subsection{Snake}
In actor-critic algorithms, such as A2C, the actor is represented by the policy network $\pi_\theta$, and the critic by the value network $v_\theta$. For the experiments on Snake, we use the A2C algorithm for control, leading to the inner loss $\mathcal{L}$ and outer loss $\mathcal{L}'$ whose gradients are derived below. The inner loss uses weights $c_{\textrm{crit}}$ and $c_{\textrm{entr}}$ that can be meta-learnt to dynamically change the impact of the critic or entropy term in the total loss.

\begin{equation*}
    \frac{\partial \mathcal{L}(\theta, \eta, \tau)}{\partial \theta} = -(g_\eta(\tau) - v_\theta(\tau))\frac{\partial \log \pi_\theta(\tau)}{\partial \theta} - c_{\textrm{crit}} (g_\eta(\tau) - v_\theta(\tau)) \frac{\partial v_\theta(\tau)}{\partial \theta} - c_{\textrm{entr}}\frac{\partial H(\pi_\theta(\tau))}{\partial \theta}
\end{equation*}

\begin{equation*}
    \frac{\partial \mathcal{L}' (\theta', \eta', \tau')}{\partial \eta} = -(g_{\eta'}(\tau') - v_{\theta'}(\tau'))\frac{\partial \log \pi_{\theta'}(\tau')}{\partial \eta}
\end{equation*}

The targets used are $\lambda$-returns from TD($\lambda$)~\cite{sutton2018reinforcement}. $g_{\eta}(\tau)$ uses the meta-parameters $\gamma$ and $\lambda$ whereas $g_{\eta'}(\tau)$ uses meta-learning hyper-parameters $\gamma'$ and $\lambda'$ for which a good proxy needs to be found. The parameters used for the experiments are described in table~\ref{tab:snake_experiments_details}.

\subsection{Hyper-Parameters}

\begin{table}[ht]
\centering
\begin{tabular}{c|c|c}
    \hline
    \textbf{Parameters} & \textbf{Markov Reward Process} & \textbf{Snake} \\ [0.5ex]
    \hline
    architecture & MLP(1,10,10,1) & conv + MLP \\
    $\gamma^{\textrm{start}}$ & 0.5 & 0.99 \\
    $\lambda^{\textrm{start}}$ & N/A  & 0.99 \\
    learning rate $\alpha$ & 1e-3 & 6e-4 \\
    learning rate schedule & constant & linear annealing to 0 \\
    inner optimiser & Adam& Adam \\
    $c_{\textrm{crit}}^{\textrm{start}}$ & N/A & 0.5 \\
    $c_{\textrm{entr}}^{\textrm{start}}$ & N/A & 0.01 \\
    gradient clipping norm & N/A & 10 \\
    batch size & 32 & 32 \\ [0.5ex]
    \hline
    $\gamma'$ & 1 & 0.99 \\
    $\lambda'$ & 1 & 0.99 \\
    meta learning rate $\beta$ & 2e-3 & 1e-3 \\
    meta optimiser & Adam & Adam \\
    meta-gradient clipping norm & N/A & 0.1 \\
    meta batch size & 32 & 32 \\ [0.5ex]
    \hline
\end{tabular}
\caption{Hyperparameters used in experiments.}
\label{tab:snake_experiments_details}
\end{table}

The architecture used for Snake is a composition of convolution layers followed by a multilayer perceptron (MLP) with two heads for the actor and the critic. The complete architecture is made of 2 consecutive blocks of one convolution layer (kernel size of 3, stride 1, and 32 feature maps) and one max pooling layer of stride and window size 2. The output of the convolution is then followed by a linear layer of output 40 from which two MLP heads of respective shape (30,4) and (30,1) compute respectively the actor and critic outputs.

We use standard values of hyper-parameters for the baseline. After meta-learning the discount factor, we yet observe that higher $\gamma$ would lead to higher performance. This shows the added value of self-tuning, in that grid searches may be avoided.

\section{Hessian Matrix}
\subsection{$n$-Step Meta-Gradient Formula}
\label{sec:n_step_meta_gradient_formula}
Here, we provide the derivation of the n-step meta-gradient:
\begin{equation*}
    \nabla_{\eta_t} \mathcal{L}' (\theta_{t+n}, \eta', \tilde{\tau}_{t+n}) = T_{t+n} \sum_{i=1}^{n} \left( \prod_{j=1}^{i-1} \left(I + H_{t+n-j}\right) \right) N_{t+n-i}
\end{equation*}

\textit{Proof.}\\
Starting from $\theta_t$, let us consider a series of $n$ inner updates, leading to $\theta_{t+n}$, through which to compute the meta-gradient $\nabla_{\eta_t} \mathcal{L}' (\theta_{t+n}, \eta', \tilde{\tau}_{t+n})$, with $\tilde{\tau}_{t+n} \sim \pi_{\theta_{t+n}}$ being validation data to compute the outer loss.
\begin{multline*}
    \forall j \in \llbracket 0, n-1 \rrbracket, \; \theta_{t+j+1} = \theta_{t+j} - \alpha \nabla_{\theta_{t+j}} \mathcal{L} (\theta_{t+j}, \eta_t, \tau_{t+j})\\
    \nabla_{\eta_t} \mathcal{L}' (\theta_{t+n}, \eta', \tilde{\tau}_{t+n}) = \underbrace{\frac{\partial \mathcal{L}'}{\partial \theta_{t+n}} (\theta_{t+n}, \eta',\tilde{\tau}_{t+n})}_{T_{t+n}} {\color{red} \frac{d \theta_{t+n}}{d \eta_t}} \\
    \theta_{t+n} = \theta_{t+n-1} \underbrace{- \alpha \nabla_{\theta_{t+n-1}} \mathcal{L} (\theta_{t+n-1}, \eta_t, \tau_{t+n-1})}_{f(\theta_{t+n-1}, \eta_t, \tau_{t+n-1})} \\
    \begin{aligned}
    {\color{red} \frac{d \theta_{t+n}}{d \eta_t}}
    &= \frac{d \theta_{t+n-1}}{d \eta_t} + \frac{\partial f(\theta_{t+n-1}, \eta_t, \tau_{t+n-1})}{\partial \eta_t} + \frac{\partial f(\theta_{t+n-1}, \eta_t, \tau_{t+n-1})}{\partial \theta_{t+n-1}} \frac{d \theta_{t+n-1}}{d \eta_t} \\
    {\color{red} \frac{d \theta_{t+n}}{d \eta_t}}
    &= \left(I + \underbrace{\frac{\partial f(\theta_{t+n-1}, \eta_t, \tau_{t+n-1})}{\partial \theta_{t+n-1}}}_{H_{t+n-1}}\right){\color{red} \frac{d \theta_{t+n-1}}{d \eta_t}} + \underbrace{\frac{\partial f(\theta_{t+n-1}, \eta_t, \tau_{t+n-1})}{\partial \eta_t}}_{N_{t+n-1}} \\
    \end{aligned} \\
    \textrm{Since} \; {\color{red} \frac{d \theta_{t}}{d \eta_t}} = 0, \quad \textrm{and} \left({\frac{d \theta_{t+i}}{d \eta_t}}\right)_{i \in \mathbb{N}} \;\; \textrm{is an arithmetico–geometric sequence}, \\
    {\color{red} \frac{d \theta_{t+n}}{d \eta_t}} = \sum_{i=1}^{n} \left( \prod_{j=1}^{i-1} \left(I + H_{t+n-j}\right) \right) N_{t+n-i} \quad \textrm{(proof by induction)}\\
    \textrm{Hence,} \quad \nabla_{\eta_t} \mathcal{L}' (\theta_{t+n}, \eta', \tilde{\tau}_{t+n}) = T_{t+n} \sum_{i=1}^{n} \left( \prod_{j=1}^{i-1} \left(I + H_{t+n-j}\right) \right) N_{t+n-i} \\
    \text{With} \begin{cases}
    T_{t+n} = \frac{\partial \mathcal{L}'}{\partial \theta_{t+n}} (\theta_{t+n}, \eta', \tilde{\tau}_{t+n}) \\
    H_{t+n-j} = \frac{\partial f}{\partial \theta_{t+n-j}}(\theta_{t+n-j}, \eta_t, \tau_{t+n-j}) = - \alpha \frac{\partial^2 \mathcal{L} }{\partial \theta_{t+n-j}^2}(\theta_{t+n-j}, \eta_t, \tau_{t+n-j}) \\
    N_{t+n-i} = \frac{\partial f}{\partial \eta_t}(\theta_{t+n-i}, \eta_t, \tau_{t+n-i}) = - \alpha \frac{\partial^2 \mathcal{L} }{\partial \theta_{t+n-i} \partial \eta_t}(\theta_{t+n-i}, \eta_t, \tau_{t+n-i}) \\
\end{cases}
\end{multline*}
The accumulative trace approximation~\cite{xu2018metagradient} consists in discarding the Hessian and using an exponential decay of previous updates parameterised by $\mu$.
\begin{equation*}
    (I + H_{t+n-j}) \approx \mu I, \quad \textrm{with} \; \mu \in [0,1]
\end{equation*}
\begin{equation*}
    \nabla_{\eta_t} \mathcal{L}' (\theta_{t+n}, \eta', \tilde{\tau}_{t+n}) \approx T \sum_{i=1}^{n} \mu^{i-1} N_{t+n-i} \qquad \text{(Accumulative trace approximation)}
\end{equation*}

\subsection{Hessian and Meta-Gradient Variance}
\label{sec:hessian_and_meta_gradient_variance}
Let us call $H_{t+n-j}$ the Hessian as it is the Hessian matrix, up to a factor $-\alpha$, of the inner loss $\mathcal{L}$ with respect to the parameters $\theta_{t+n-j}$ of the network. It is of size the square of the number of parameters, which can quickly become huge. Additionally, we presume the meta-gradient variance to partly come from this Hessian. Indeed, the $n$-step meta-gradient estimation uses products of these Hessian matrices, and multiplying matrices may lead to high matrix norm, escalating variance.

Let us denote by $P_i$ one such product $\prod_{j=1}^{i-1} \left(I + H_{t+n-j}\right)$. In the worst case, all Hessian matrices are equal and hence, the meta-gradient norm may explode. Therefore, to study this impact on the variance, we assume a common Hessian matrix $H$ for the $n$ steps, $H_{t+n-j} \approx H$ symmetric as well.

Then, $\left(I + H\right)$ is a real and symmetric matrix, hence the spectral theorem~\cite{douglas_pearcy_1970} states that there exists an orthogonal matrix $Q$ and a diagonal matrix $D$ such that $\left(I + H\right) = QDQ^{-1}$.

\begin{equation}
    P_i = \prod_{j=1}^{i-1} \left(I + H_{t+n-j}\right) \approx \prod_{j=1}^{i-1} \left(I + H\right) = QD^{i-1}Q^{-1}
\end{equation}

\begin{equation}
    \lambda \in Sp(H) \iff \lambda + 1 \in Sp\left(I + H\right) \iff (\lambda + 1)^{i-1} \in Sp\left(P_i\right)
\end{equation}

Therefore, an eigenvalue of the Hessian that lies outside the interval $[-2,0]$ is to explode in the product $P_i$ for large $i$, leading to a high spectral radius and a high matrix norm. Meaning that the randomness caused by differentiating the inner update computed in $N_{t+n-i}$ is likely to be amplified by high-norm $P_i$, leading to high meta-gradient variance.



\end{document}